\documentclass[letterpaper, paper,11pt]{AAS}		% for final proceedings (20-page limit)

\usepackage{bm}
\usepackage{amsmath}
\usepackage{subfigure}
\usepackage[colorlinks=true, pdfstartview=FitV, linkcolor=black, citecolor= black, urlcolor= black]{hyperref}
\usepackage{overcite}
\usepackage{footnpag}			      	% make footnote symbols restart on each page

\usepackage[utf8]{inputenc}
\usepackage{algorithm}
\usepackage{algpseudocode}

\usepackage{graphicx, float, listings, hyperref}
\usepackage{cite}
\usepackage{amsmath, amssymb, bm}

\usepackage{multirow}
\usepackage{caption}

\usepackage{siunitx}
\PaperNumber{22-705}

\begin{document}

\title{ TOWARDS A GENERALIZABLE SIMULATION FRAMEWORK TO STUDY COLLISIONS BETWEEN SPACECRAFT AND DEBRIS
%MANUSCRIPT TITLE (UP TO 6 INCHES IN WIDTH AND CENTERED, 14 POINT BOLD FONT, MAJUSCULE)
}

\author{Simone Asci\thanks{Ph.D. Candidate, School of Engineering and Materials Science, Queen Mary University of London, London, E1 4NS, UK, s.asci@qmul.ac.uk},  
Angadh Nanjangud\thanks{Lecturer in Spacecraft Engineering, School of Engineering and Materials Science, Queen Mary University of London, London, E1 4NS, UK, a.nanjangud@qmul.ac.uk}
}

\maketitle{}

\begin{abstract}
In recent years, computer simulators of rigid-body systems have been successfully used to improve and expand the field of developing new space robots, becoming a leading tool for the preliminary investigation and evaluation of space robotic missions. However, the impressive progress in performance has not been matched yet by an improvement in modelling capabilities, which remain limited to very basic representations of real systems. We present a new approach to modelling and simulation of collision-inclusive multibody dynamics by leveraging symbolic models generated by a computer algebra system (CAS). While similar investigations into contact dynamics on other domains exploit pre-existing models of common multibody systems (e.g., industrial robot arms, humanoids, and wheeled robots), our focus is on allowing researchers to develop models of novel designs of systems that are not as common or yet to be fabricated: e.g., small spacecraft manipulators. In this paper, we demonstrate the usefulness of our approach to investigate spacecraft-debris collision dynamics.
\end{abstract}

\label{ch:Intro}
\section{Introduction}

Computer simulations of collision/contact dynamics is a germane research topic of engineering science, especially within the mechanics \cite{dojo} and robotics communities.\cite{mylapilli} It is also one of great importance in various space mission scenarios.\cite{papadopoulos2021robotic, nanjangud2018} For example, in-orbit assembly and servicing missions require an understanding of interactions between bodies that make and break contact with each other during proximity operations (e.g., rendezvous and docking of space vehicles; in-orbit assembly operations such as manipulation by and relocation of robotic agents\cite{nanjangud2019}), which can most affordably and speedily be achieved via computer-aided engineering. Contact simulations are also useful in providing insight into the motion behavior of satellites during and after collisions with space debris, the focus of this paper.
The aforementioned research on computational contact dynamics tackles improving real-time simulation accuracy and speed, which focuses on the underlying numerical optimisation routines. While important strides have been made here, much of this research exploits predefined models of multibody systems (e.g., 6-DOF manipulators, humanoids, wheeled robots, quadrupeds). As a result, the research tools do not generalise to other domains where contact studies on more niche physical systems are necessary. For example, advances in contact simulations on humanoids are not directly applicable to spacecraft. This lack of a generalizable simulation framework is a key concern within the larger robotics community.\cite{pnas} Thus, while our main focus is on modelling and simulation of collisions between spacecraft and debris, our developments also consider tackling this larger problem by of interdisciplinary modeling of multibody systems with contact.
Here, we develop a contact dynamics model describing the behavior of a system using symbolic equations of motion generated by a computer algebra system (CAS). A major advantage here is that the dynamics of two (or more) interacting systems is completely defined and accessible to a modeller in symbolic form prior to numerical simulations. This is in contrast to current approaches, where models are treated as black boxes; this may also further explain why most recent work on contact dynamics has focused on commercially available robotic systems (e.g., humanoid\cite{dojo} and quadrupeds\cite{manchester2020variational} manufactured by Boston Dynamics). We present results from a variety of experiments ranging from simple (a ball bouncing on a surface) to complex (collisions between a spacecraft and debris). On account of exposing the underlying dynamics models as symbolic functions, our approach introduces a degree of flexibility in implementing contacts, allowing for a wide variety of contact models.\cite{papadopoulos2021robotic, contactreview} In our work, we exploit an elasto-plastic contact model, previously used in space manipulator research.\cite{rybus}
Such collision-inclusive dynamics models can play a fundamental role in the field of spacecraft dynamics and control. First, they provide a means to reproduce and analyze the uncontrolled motion caused by an on-orbit impact, which is governed by contact dynamics and represents an active area of research.\cite{papadopoulos2021robotic} Second, the spacecraft's response to collisions then informs appropriate control algorithm development for re-stabilizing its motion.\cite{mote2020collision} Last but not least, such models will also have bearing in future developments on autonomous on-orbit robotic missions (e.g., rendezvous and berthing, assembly, servicing, and refueling) in areas such as task and motion planning.\cite{moserFrontiers2022}

\label{ch:Problem}
\section{System description}

This paper deals with dynamics modelling and simulation of a two-link manipulator mounted on a base spacecraft that experiences an impact with debris at some point during an operation of target capturing. Specifically, it addresses the problem of emulating the uncontrolled motion caused by the impact and informs on how this affects the spacecraft controller's capability to generate a recovery maneuver and complete the original task.

\begin{figure}[htb]
	\centering\includegraphics[width=3.5in]{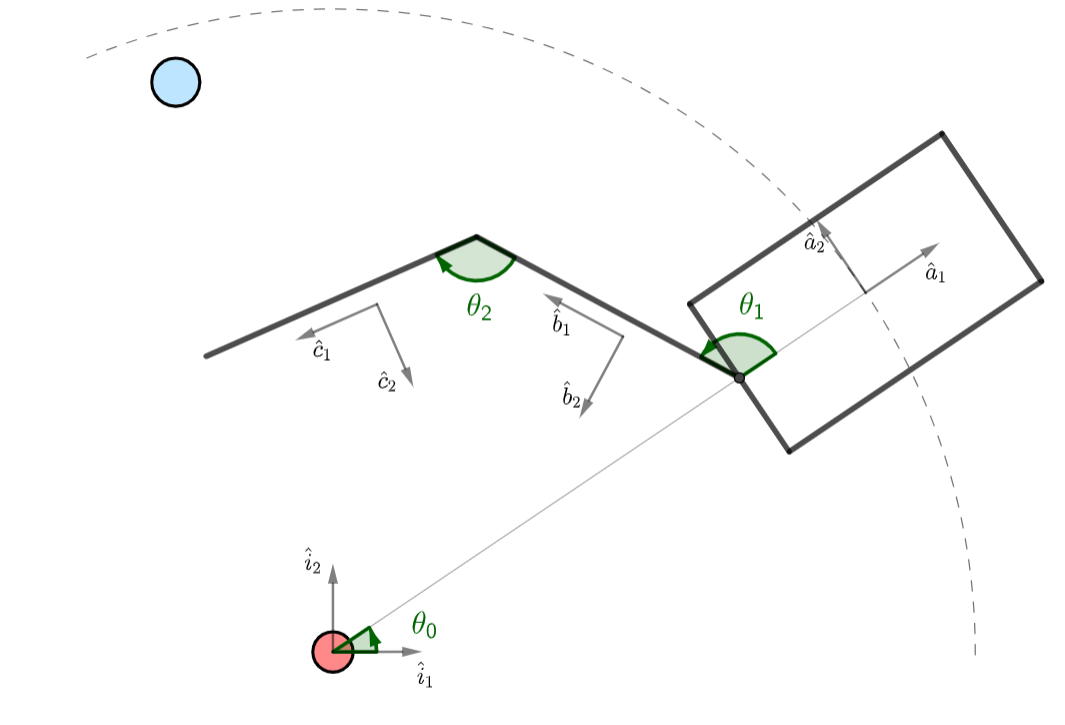}
	\caption{ \textbf{Spacecraft and Debris Systems}}
	\label{fig:endeffec}
\end{figure}

\noindent
With reference to Figure~\ref{fig:endeffec}, a space target (highlighted in red) is supposed to orbit around the Earth. A spacecraft flies freely around the target on a circular path with a constant angular velocity. The 2-links manipulator attached to the base is programmed to execute a target-capturing maneuver, starting from a folded configuration and following a parabolic trajectory. The end-effector (EE) should approach the target with zero relative velocity at the touching moment, in order to meet the requirements of soft-touch.
At some point during the maneuver execution, the manipulator experiences a collision with debris (in blue) which causes a loss in the prescribed trajectory tracking, whose severity 
depends on the collision direction and magnitude.
Upon the impact event, the onboard controller starts a compensating maneuver aimed to stabilize the error to zero. The main factors in assessing the effectiveness of the controller are the recovery time of the original trajectory and motor torques. 
The simulation is based on the assumption that the motion of the base with respect to the target is out of consideration for dynamics modelling,\cite{kane2016} being not influenced by the manipulator dynamics and collisions; furthermore, all the bodies considered are characterized by coplanar motions. 
A collision-inclusive dynamics model of the spacecraft is developed in symbolic form using the proposed simulation framework. It is realised with a flexibility and configurability that leaves the modeller the arbitrary choice of the multi-body system and contact model to be applied, while maintaining a modular and procedural implementation order that allows easy extension to different domains.

\label{ch:Framework}
\section{Simulation Framework}

%SHORT DESCRIPTION OF THE PARAGRAPH
%  Here I describe how to model in a systematic way any given dynamic system through the symbolic generation of EOMs.
%Then I explain how contact is modelled in a general way by using placeholders which can be replaced in numerical simulations with arbitrary values.
% I briefly compare the "total symbolic approach " with the "hybrid" one.

The proposed simulation framework is based on SymPy, a computer algebra system (CAS) implemented in Python, which allows the performing of algebraic operations with symbolic expressions.\cite{pendulumconstrained} A fully symbolic expression offers several advantages over its numerical counterpart, since it is characterised by ease of formulation, replication and manipulation, this guarantees substantial improvements in facilitating the examination and interpretability of results. Furthermore, SymPy offers unique flexibility in the interchangeability of expressions between their symbolic and numerical form, which results in an advantageous trade-off between modelling power and computational speed.
The modular design of the proposed framework is composed of two pieces. The first is the symbolic modelling of the dynamic system, completely decoupled from the contact model, which is implemented symbolically as well. The second piece is the dynamics system simulation, through the numerical evaluation of the symbolic equations of motion.

\subsection{Symbolic Module}
The symbolic modelling of a dynamic system consists of a sequence of modular steps to obtain the equations of motion (EoMs) in symbolic form. In this work, SymPy tool is utilized, in addition to a mechanics toolkit optimized for modelling multibody systems with many degrees of freedom. SymPy integrates with other Python libraries for numerical calculus, allowing for an automatic transition from the great representative power of the symbolic world to the computational power of the numeric world, furthermore, it employs simplification routines which often allow for evaluating EoMs faster than their numerical counterparts, with consequent advantages for real-time simulations.\cite{pendulumconstrained}
The most important advantage provided by this method is the possibility to derive the EoMs in an automated fashion, a task that is challenging and slow to perform by hand for complex multibody systems, and it is very difficult and prone to error even with the support of a CAS.

\subsection{Numeric Module}

The second piece of the framework modular design consists of the numeric simulation of the dynamic system. Symbolic EoMs are converted to their numerical equivalent expressions and integrated with the numerically evaluated expressions of contact dynamics; thus the evolution of the system in function of time can be obtained through numerical integration.

\label{ch:Contact}
\section{Contact Model}

% Then I specifically explain the contact model we use for simulations, with the following steps:
%1) contact detection based on the SAT (separating axle theorem)
%2) computation of force direction.
%3) force magnitude based on Rybuzs formulas

% Then I add 2 pseudocodes, one for a circular debris and another for a polygonal shape debris.

Contact is a very complex phenomenon and remains a key challenge area in the field of dynamics since it is highly non-linear and hard to model;\cite{papadopoulos2021robotic} the majority of proposed approaches relies on case-specific models, designed with a limited scope of validity and applicable only in determined conditions. More research work is needed on this topic as in the open literature there is not a general-purpose systematic approach to formulate impacts dynamics in generalized scenarios.\cite{dojo} 
The general procedure to compute the contact between interacting bodies consists of two steps:
\vspace{-\topsep}
\begin{enumerate}
     \setlength{\parskip}{0pt}
    \item collision detection, to determine the overlap between the bodies
    \item collision resolution, consisting in resolving the direction and magnitude of interacting forces 
\end{enumerate}
\vspace{-\topsep}

\noindent
Collision detection is a computational geometry problem with applications in various fields, including computer graphics and video games. Popular approaches are those based on Minkowsky difference, like the GJK algorithm,\cite{gjk} and those based on the Separating Axis Theorem (SAT),\cite{separating} like the Axis-Aligned Bounding Box.\cite{aabb} Besides traditional methods, advanced techniques have been proposed as well to predict collision-free configurations, such as the Dynamic Collision Checking which is based on heuristic search.\cite{adaptivecollision} The algorithm choice relies on assessing the trade-off between detection accuracy and computational effort, since collision detection is the main responsible for simulation slow down.\cite{pnas}  
The computation of collision force is a problem related to physics and mechanics. It depends on several factors, primarily the materials and mechanical properties of the colliding bodies; since it is still not fully comprehended, numerous models proposed in the literature are often based on empirical observations and experimental data.\cite{mote2020collision}
These methods can be classified in two categories: the discrete and continuous models.\cite{wu2017contact} The discrete approach resolves contact only for rigid bodies based on kinematics constraints. 
The continuous approach models the force based on the local deformation of the contact bodies, computed as the intersection between the respective geometries; it has been widely applied in robotic contact problems, due to its suitability in handling complex geometries. It comprehends the category of compliance-based models, including the bristle-friction and the elastic-plastic.
The Bristle friction model is based on the linear approximation of the Coulomb friction model.\cite{bristle}
The Elastic-plastic model, models the interacting force as the sum of two contributions depending on the bodies' relative velocity and the amount of local deformation.\cite{rybus}

\subsection{Force Magnitude Computation}

In this work, the magnitude of each force component is computed based on the Elastic-plastic approach, which designs the interaction between colliding bodies as a spring-damping model. Complete information about the material, geometry, and velocity of both the bodies involved is supposed to be known.
With reference to Eq.~\eqref{eq:1} the normal force is composed of the elastic component, proportional to the distance between centers of mass, and the plastic component which depends on the relative velocity along the normal direction.\cite{rybus} The tangent force depends on the friction between the surfaces and the relative velocity along the tangent direction.

\begin{equation}
\begin{cases}
F_{N} =   (kc \times p^{3}) \times (1 -cc \times v_{N} )  \\
F_{T} =  -mu \times F_{N} \times  ( \frac{2}{ 1 + e^{ \frac{-v{T}}{vs}}  }  -1 )
\end{cases}
\label{eq:1}
\end{equation}

\noindent
Where $p$ is the interpenetration, i.e. the local indentation of the contact surface, $kc$, and $cc$ are the contact stiffness parameter and damping coefficient respectively, $mu$ is the friction coefficient and $vs$ is a scaling factor, $v_{N}$ and $v_{T}$ are the normal and tangent values of the relative velocity.\cite{ackermann2010optimality}

\label{ch:Spacecraft}
\section{Dynamics modelling and control study for spacecraft and debris system}

% Here I describe the Spacecraft model and the debris, with respective EOMs and how collision happen ( e.g. only one-point planar collision...).
% I add the control model.

% To structure this chapter we can generally follow this paper: "Kane Method Based Dynamics Modeling and Control Study for Space Manipulator Capturing a Space Target"

The examined scenario comprises two interacting bodies. The first one is a base spacecraft with an attached manipulator; it is modelled as a multibody system with constant orbital motion around a target, which is represented as a simple marker point. The second one is a debris, modelled as a single rigid body.
The analyzed motion is restricted to a plane and subjected to the microgravity conditions of the orbital environment.

%\begin{figure}[h]
%	\centering
%	\includegraphics[width=0.9\columnwidth]{Pictures/vehicle.PNG}
%	\caption{Spacecraft and debris systems} \label{fig:endeffec}
%\end{figure}

\subsection{Preliminary Work}
A space target orbiting around the Earth is supposed to be fixed in an inertial reference system. A base spacecraft flies around the target on a circular orbit of radius $R$, in the anticlockwise direction. With reference to figure \ref{fig:endeffec}, the right-handed inertial frame is expressed as $O_Ix_Iy_I$, the origin $O_I$ is coincident with the position of the target, axis $x_I$ points an arbitrary fixed direction, axis $y_I$ is got by rotating  with 90° anticlockwise. Unit vectors $\textbf{I}=(\hat{i}_1, \hat{i}_2)$ lie on $x_I$ and $y_I$ respectively with the same directions.
The manipulator links are described as rigid, homogenous bodies, their mass and moment of inertia with respect to their centers of mass are $m_i$ and $J_i$, while their lengths are identified as $l_i$ ($i = 1,2$), these parameters are equal for both the links. The first joint, indicated as I, is mounted on the base larger side, at a distance $d$ from the base center of mass, and it is driven by torque $M_1$; the second joint and relative torque are indicated as II and $M_2$, respectively. The debris is instead modelled as a particle, characterized only by a mass $m_d$ and a radius $r_d$.
Consistently with the notation used for the inertial frame, local frames are defined for each body as follows:

\begin{itemize}
    \item spacecraft base local frame $O_Ax_Ay_A$, the origin $O_A$ is at the center of mass, axis $x_A$ lies on the line from $O_I$ to $O_A$ with a positive direction deviating from $O_I$, while axis $y_A$ is obtained by rotating $x_A$ with 90° anticlockwise. Unit vectors $\textbf{A}=(\hat{a}_1, \hat{a}_2)$ lie on respective axes with the same directions. 
    \item link I local frame $O_Bx_By_B$, the origin $O_B$ is at the center of mass, with $x_B$ axis pointing towards joint II and $y_B$ axis obtained as for the base local frame. Unit vectors $\textbf{B}=(\hat{b}_1, \hat{b}_2)$ lie on respective axes with the same directions. 
    \item link II local frame $O_Cx_Cy_C$, the origin $O_C$ is at the center of mass, with $x_C$ axis pointing towards the end-effector and $y_C$ axis obtained as for the base local frame. Unit vectors $\textbf{C}=(\hat{c}_1, \hat{c}_2)$ lie on respective axes with the same directions.  
\end{itemize}

\noindent
The orientation of the spacecraft is determined by the angle $\theta_0$, which indicates the rotation of $O_Ax_Ay_A$ with respect to $O_Ix_Iy_I$, positive in the anticlockwise direction. Orientations of $O_Bx_By_B$ with respect to $O_Ax_Ay_A$ and of $O_Cx_Cy_C$ with respect to $O_Bx_By_B$ are indicated with $\theta_1$ and $\theta_2$ respectively. The transformations relating unit vectors are defined in Eq.~\eqref{eq:2}:

%\begin{flalign}
%& \mathbf{A} = \mathbf{{R}_{IA}}  \mathbf{I}\\
%& \mathbf{B} = \mathbf{{R}_{AB}}  \mathbf{A}  =  \mathbf{{R}_{IB}} \mathbf{I}\\
%& \mathbf{C} = \mathbf{{R}_{BC}}  \mathbf{B}  =  \mathbf{{R}_{IC}} \mathbf{I}
%\end{flalign}
%\label{eq:2}

\begin{equation}
\begin{split}
& \mathbf{A} = \mathbf{{}^{I}{R}^{A}} \cdot \mathbf{I}\\
& \mathbf{B} = \mathbf{{}^{A}{R}^{B}} \cdot \mathbf{A}  =  \mathbf{{}^{I}{R}^{B}}\cdot \mathbf{I}\\
& \mathbf{C} = \mathbf{{}^{B}{R}^{C}} \cdot \mathbf{B}  =  \mathbf{{}^{I}{R}^{C}} \cdot \mathbf{I}
\end{split}
\label{eq:2}
\end{equation}

\noindent
where $\mathbf{{}^{I}{R}^{A}}, \mathbf{{}^{A}{R}^{B}}, \mathbf{{}^{B}{R}^{C}}  \in \mathbb{R}^{3\times3}$ are the rotation matrices relating the world frame $O_Ix_Iy_I$ to frame $O_Ax_Ay_A$, frame $O_Ax_Ay_A$ to $O_Bx_By_B$ and $O_Bx_By_B$ to $O_Cx_Cy_C$, respectively.   
They are defined as follows:

\begin{equation}
\begin{split}
	&\mathbf{{}^{I}{R}^{A}}= 
    \begin{pmatrix}
      \cos\theta_0  & \sin\theta_0   & 0 \\
    - \sin\theta_0  &  \cos\theta_0   & 0 \\
      0              &      0           & 1
    \end{pmatrix} \\   
    &\mathbf{{}^{A}{R}^{B}}=
    \begin{pmatrix}
      \cos\theta_1  & \sin\theta_1  \\
    - \sin\theta_1  & \cos\theta_1   
    \end{pmatrix} \\
    &\mathbf{{}^{B}{R}^{C}}= 
    \begin{pmatrix}
     - \cos\theta_2  &  \sin\theta_2  \\
     - \sin\theta_2  &  - \cos\theta_2    
    \end{pmatrix} \\
    & \mathbf{{}^{I}{R}^{B}} = \mathbf{{}^{I}{R}^{A}}  \cdot \mathbf{{}^{A}{R}^{B}}  \\
    & \mathbf{{}^{I}{R}^{C}} = \mathbf{{}^{I}{R}^{A}} \cdot \mathbf{{}^{A}{R}^{B}} \cdot \mathbf{{}^{B}{R}^{C}} \\
  	\label{eq:3}
\end{split}  	
\end{equation}

\noindent
The position, velocity, and acceleration of the base center of mass with respect to the inertial frame are: 

\begin{equation}
	\vec{r}_{a} =  
	\mathbf{A} \begin{pmatrix}
     R \\ 0  \end{pmatrix}, 
    \quad \vec{v}_{a} =  \frac{d\vec{r}_{a}}{dt},
    \quad \vec{a}_{a} =  \frac{d^2\vec{r}_{a}}{dt^2}
\label{eq:4}
\end{equation}

\noindent
Applying direct kinematics and then taking the first and second order time  derivatives yield the kinematics equations of the relevant points. For the first link center of mass: 

\begin{equation}
	\vec{r}_{b} =  \vec{r}_{a} + 
	\mathbf{B} \begin{pmatrix}
     -d\cos\theta_1 + \frac{l_1}{2}  \\ d\sin\theta_1  \end{pmatrix}, 
    \quad \vec{v}_{b} =  \frac{d\vec{r}_{b}}{dt},
    \quad \vec{a}_{b} =  \frac{d^2\vec{r}_{b}}{dt^2}
\label{eq:5}
\end{equation}

\noindent
For the second link: 

\begin{equation}
	\vec{r}_{c} =  \vec{r}_{b} + 
	\mathbf{C} \begin{pmatrix}
     \frac{l_1}{2}\cos\theta_2 + \frac{l_2}{2}  \\ -\frac{l_1}{2}\sin\theta_2  \end{pmatrix},
    \quad \vec{v}_{c} =  \frac{d\vec{r}_{c}}{dt},
    \quad \vec{a}_{c} =  \frac{d^2\vec{r}_{c}}{dt^2}
\label{eq:6}
\end{equation}

\noindent
Finally, for the end-effector: 

\begin{equation}
	\vec{r}_{EE} =  \vec{r}_{c} + 
	\mathbf{C} \begin{pmatrix}
     \frac{l_2}{2}  \\ 0  \end{pmatrix},
    \quad \vec{v}_{EE} =  \frac{d\vec{r}_{EE}}{dt},
    \quad \vec{a}_{EE} =  \frac{d^2\vec{r}_{EE}}{dt^2}
\label{eq:7}
\end{equation}

\noindent
From the first of Eq.~\eqref{eq:7}, the position of the end-effector expressed in the inertial frame is:

\begin{equation}
	\vec{r}_{EE} =  
	\mathbf{I} \begin{pmatrix}
     (R-d)\cos\theta_0 + l_1\cos(\theta_0+\theta_1) + l_2\cos(\theta_0+\theta_1+\theta_2) \\
     (R-d)\sin\theta_0 + l_1\sin(\theta_0+\theta_1) + l_2\sin(\theta_0+\theta_1+\theta_2)    
    \end{pmatrix} \\
\label{eq:8}
\end{equation}

\noindent
The debris is not provided with a local reference frame, its position is defined as:
\begin{equation}
	\vec{r}_{d} =  
	\mathbf{I} \begin{pmatrix}
    x_{d} \\  y_{d}  \end{pmatrix},
    \quad \vec{v}_{d} =  \frac{d\vec{r}_{d}}{dt},
    \quad \vec{a}_{d} =  \frac{d^2\vec{r}_{d}}{dt^2}
\label{eq:9}
\end{equation}

\subsection{Dynamics Model of the System  }
The base spacecraft motion is considered to be not influenced by any external force, including manipulator-coupled dynamics and eventual collisions with the debris, based on the assumption that its mass and inertia are far larger than those of other bodies. Furthermore, it is assumed that the base is facing always the space target in the same orientation during the entire simulation, a requirement that can be satisfied with an accurate attitude control system.    
Conversely, we suppose that both the base relative movement and eventual collisions have an active role in the 2-link manipulator dynamics modelling.
Despite the target flying motion around the earth should be defined in a non-inertial coordinate frame, its translational motion can be out of consideration, since it is caused by gravity which can be ignored as well.\cite{kane2016} The angular rotation around its own origin should instead be taken into account for the dynamic modeling of the system. However, this rotation is performed at the same angular velocity of the target revolution around Earth, which is very low. Since the analysis in this paper is concerned with a short time interval compared to the target orbital period, its angular rotation can be neglected, and therefore we can suppose that the target is fixed in the inertial reference frame, with both translational and rotational motions equal to zero.
Furthermore, all the centrifugal forces in the target inertial frame counteract exactly the gravitational ones and so can be neglected, which therefore allows  any effect caused by the target orbit dynamics to be disregarded.\cite{kane2016} Consequently, only the base spacecraft relative orbit dynamics is taken into account in the system dynamics modeling; the spacecraft coordinate system $O_Ax_Ay_A$ moves around the target and rotates about its own origin at the same angular velocity ($\dot{\theta_0}$), in this way the hypothesis of constantly pointing the target with the same side is satisfied. 
The dynamics equations of the manipulator can be derived in terms of generalized coordinates as shown in Eq.~\eqref{eq:10}:

\begin{equation}
	B(\theta) \ddot{\theta} + c(\theta,\dot{\theta}) + g(\theta)  = \tau + F_c
\label{eq:10}
\end{equation}

\noindent
where $\theta \in R^{n}$ represents the generalized coordinates of the spacecraft, with $n$ equal to the number of links (2 in this case), $B(\theta)$ is the inertia matrix of the manipulator, $c(\theta,\dot{\theta})$ expresses the nonlinear Coriolis and centrifugal terms, $g(\theta)$ expresses the potential terms, $\tau \in R^{n}$  represents joint torques, and finally $F_c$ is the collision forces vector.
The debris is subjected to the interaction forces caused by collisions, and to a constant attractive force that drives it towards the spacecraft.

\subsection{Trajectory Planning}

The manipulator trajectory and relative control function are programmed so that the end-effector approaches the target executing a parabolic motion, with the final direction perpendicular to the orbit radius and final velocity equal to zero, so as to meet the soft-touch requirements. 
At initial moment, the manipulator is folded at $\theta_1$ = $\theta_{10}$ and $\theta_2$ = $\theta_{20}$ where $\theta_{10}$ and $\theta_{20}$ are small values. 
Data relative to the trajectory is: 
\begin{itemize}
    \item initial point $\vec{p}_i$, correspondent to the initial end-effector position
    \item final point $\vec{p}_f$, the target position in the inertial frame, coincident with the origin $O_I$
    \item final trajectory direction $\vec{d}_f$: perpendicular to the line connecting $O_I$ and $O_A$ 
    \item initial moment $t_i$, equal to 0 
    \item final moment $t_f$, when the EE touches the target
    \item initial EE velocity $v_i$, equal to 0 
    \item final EE velocity $v_f$, equal to 0 
\end{itemize}

\noindent
The parabolic maneuver is generated using cartesian trajectory planning with relevant constraints on initial and final positions and velocities. Using the separation in space and time, it will firstly be computed the parametric cartesian path expression $\vec{p}(s)$ and then the associated timing law $s(t)$. Given the general quadratic function for the path $\vec{p}(s)$ and its first and second derivatives relative to $s$:
\begin{equation}
\begin{split}
    &\vec{p}(s) = \vec{a} + \vec{b}s + \vec{c}s^2,  \qquad  s \in [0,1] \\
    &\vec{p'}(s)= \vec{b} + 2\vec{c}s, \\
    &\vec{p''}(s)= 2\vec{c}
\end{split}   
\label{eq:11}
\end{equation}

\noindent
and imposing the boundary conditions:
\begin{equation}
    \vec{p}(0)=\vec{p}_i, \qquad  \vec{p}(1)=\vec{p}_f,  \qquad  \vec{p^{\prime}}(1)=\vec{d}_f 
   \label{eq:12}
\end{equation}

\noindent
The parameters for $\vec{p}(s)$ can be obtained according to Eq.~\eqref{eq:11} and Eq.~\eqref{eq:12}:
\begin{equation}
    \vec{a}=\vec{p}_i, \qquad  \vec{b}=2(\vec{p}_f-\vec{p}_i)-\vec{d}_f,  \qquad  \vec{c}=\vec{d}_f - (\vec{p}_f-\vec{p}_i) 
   \label{eq:13}
\end{equation}

\noindent
The timing law and its derivatives are given by the following polynomial functions:
\begin{equation}
\begin{split}
    &s(t) = c_0 + c_1 t + c_2 t^2 + c_3t^3,  \qquad  t \in [0,t_f] \\
    &\dot{s}(t)= c_1  + 2c_2t + 3c_3t^2, \\
    &\ddot{s}(t)= 2c_2 + 6c_3t
\end{split}  
\label{eq:14}
\end{equation}

\noindent
with boundary conditions defined in Eq.~\eqref{eq:15}:
\begin{equation}
    s(0)=0 , \qquad  s(t_f)=1,  \qquad  \dot{s}(0)=0, \qquad  \dot{s}(t_f)=0 \label{eq:15}
\end{equation}

\noindent
Imposing Eq.~\eqref{eq:15} in Eq.~\eqref{eq:14} and solving, yields these values for the parameters of $s(t)$:  
\begin{equation}
    c_0=0 , \qquad  c_1=0,  \qquad  c_2=3\left( \frac{1}{t_f} \right)^2, \qquad  c_3=2 \left( \frac{1}{t_f} \right)^3 
   \label{eq:16}
\end{equation}

\noindent
Finally, from the separate profiles in space and time defined in Eq.~\eqref{eq:11} and Eq.~\eqref{eq:14} respectively, we can recombine the trajectory:
\begin{equation}
    \vec{p}(t)=\vec{p}(s(t)),  \quad   \dot{\vec{p}}(t)=\vec{p^{\prime}}(s(t))\dot{s}(t),  \quad     \ddot{\vec{p}}(t)=\vec{p^{\prime}}(s(t))\ddot{s}(t) +  \vec{p^{\prime\prime}}(s(t))\dot{s}^2(t),   \quad  t \in [0,t_f] 
   \label{eq:18}
\end{equation}

\noindent
This twice-differentiable cartesian trajectory is translated to correspondent joint trajectories through inverse kinematics, which allows for the detection of any singularity.

\subsection{Design of Control Torques}
A general control technique applied in the robotics field for trajectory tracking is composed of a feedforward component to achieve the nominal required torques and a feedback component to minimize position and velocity tracking errors of the end-effector. 
Since in this scenario both the manipulator dynamic model and a twice-differentiable desired trajectory are available, the ``computed torque controller" represents an appropriate choice to generate the torque functions for trajectory execution. In fact, it allows exploiting its feedforward and feedback components to compensate for non-linearities and stabilize the trajectory error to zero. The formula is shown in Eq.~\eqref{eq:19}:
\begin{equation}
	\tau = B(\theta) [\ddot{\theta}_d + K_p(\theta_d - \theta) + K_d(\dot{\theta}_d - \dot{\theta}) ] + c(\theta,\dot{\theta}) + g(\theta)  
\label{eq:19}
\end{equation}

\noindent
where the desired acceleration $\ddot{\theta}_d$ is obtained from trajectory planning, $K_p$ and $K_d$ are positive definite matrices whose gains are set by trial, $B(\theta)$, $c(\theta,\dot{\theta})$, $g(\theta)$ terms are directly accessible inspecting the manipulator dynamic model.
It should be considered, however, that all the effects introduced by collision forces can not be foreseen by the control, and as a result of an impact, the manipulator will experience a deviation from its trajectory. The ability to recover the original motion depends on the direction and magnitude of applied forces and will be the focus of this paper's simulations.

\label{ch:Simulations}
\section{Numerical Simulations}

% The 2 main examples can be :
% 1) bouncing ball, which highlights the effectiveness of contact model
% 2) Collision between Spacecraft and circular debris, which highlights more the reproduction of uncontrollable motion and re-stabilization through control

%% Then we can add other scenarios, like:
%- 2 balls
%- ball - rectangle

%% But I think it is more interesting to do further simulations with the spacecraft and how reacts to impact in following conditions:
%- fixed base
%- free flying
%- free floating
%- exploiting the redundancy...

A preliminary test of the contact model reliability is performed on the simple scenario of a bouncing ball. Simultaneously, an overall validation of the proposed symbolic framework is executed, analyzing its modelling accuracy and execution time in comparison with an identical simulation based on the traditional numeric method. 
Subsequently, the collision between the base spacecraft and debris will be analyzed.

\subsection{Bouncing Ball}
This case involves collisions between simple-shaped rigid bodies, the EoMs are characterized by a limited number of variables, and forces are expressed with very simple formulas. Therefore they are effectively implemented with both symbolic and numerical methods, and there is not an appreciable difference in model complexity or execution time.

\begin{figure}[!hb]
	\centering\includegraphics[width=3.5 in]{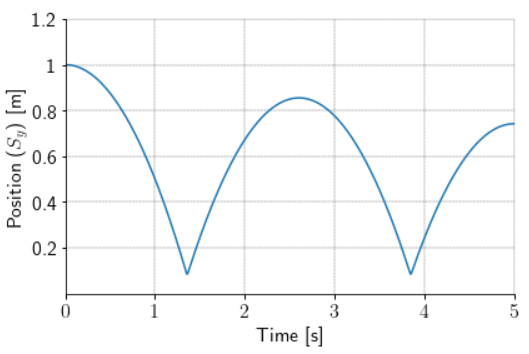}
	\caption{\textbf{Post-Collision Planar Displacement of 2-D Bouncing Ball}}
	\label{fig:ball}
\end{figure}

\noindent
Figure~\ref{fig:ball} shows the trajectory of a stationary particle (the ball) being released from 1 meter altitude and subjected to a downward force of 1 N until it hits the ground, modelled as a simple horizontal line. The force experienced by the particle during the collision causes an update of its motion in the opposite direction to the impact. The particle subsequently experiences another collision of decreasing magnitude and responds in a similar way. This solution replicates the realistic behavior that can be appreciated in a ball bouncing on the ground, and therefore it demonstrates the fidelity of the model for this simple case.

\vspace{10 pt}

\subsection{Collision Between Spacecraft and Debris}
This scenario highlights the reproduction of an uncontrolled motion caused by a collision between a spacecraft and debris and the re-stabilization capability of the adopted control law. It provides a demonstration of the framework's ability to simulate a spacecraft trajectory in complex conditions. The parameters used are shown in Table~\ref{tab:1}:

\def\arraystretch{1.2}
\def\tabcolsep{16pt}
\begin{table}[htbp]
	\fontsize{10}{10}\selectfont
    \caption{\textbf{Parameters for Numerical Simulation}}
   \label{tab:1} 
        \centering 
   \begin{tabular}{l  | r  } % Column formatting, 
      \hline 
      Parameter    & Value   \\
      \hline
      $m_1$ (\si{kg})                  & 2       \\   
      $m_2$ (\si{kg})                  & 2       \\
      $l_1$ (\si{m})                   & 3       \\
      $l_2$ (\si{m})                   & 3       \\
      $d$ (\si{m})                     & 2     \\
      $J_{1}$ (\si{kg.m^{2}})         & 6       \\
      $J_{2}$ (\si{kg.m^{2}})         & 6       \\
      $m_d$ (\si{kg})                  & 1       \\
      $r_d$ (\si{m})                   & 0.1       \\
      $R$ (\si{m})                     & 4        \\      
      $x_0$ (\si{m})                   & 4       \\
      $y_0$ (\si{m})                   & 0       \\
      $\theta_{10}$ (\si{rad})         & 0.79       \\
      $\theta_{20}$ (\si{rad})         & 1.31       \\      
      $\omega_0$ (\si{rad/s})          & 0.1       \\     
      $x_{d0}$ (\si{m})                & 7       \\
      $y_{d0}$ (\si{m})                & 5       \\
      $t_f$ (\si{s})                   & 20       \\
      \hline
   \end{tabular}
\end{table}

\noindent
The manipulator links start at rest position, while the debris is driven by an attractive force of 1 N towards the center of mass of link II; causing one collision during the simulation interval. Overall, the numerical simulation results demonstrate the effectiveness of the control law in the trajectory tracking task, in fact the actual trajectory of the end-effector almost overlaps the prescribed trajectory exactly, except for the instant corresponding to collisions.

\begin{figure}[!h]
  \centering
  \begin{minipage}[b]{0.4\textwidth}
    \includegraphics[width=\textwidth]{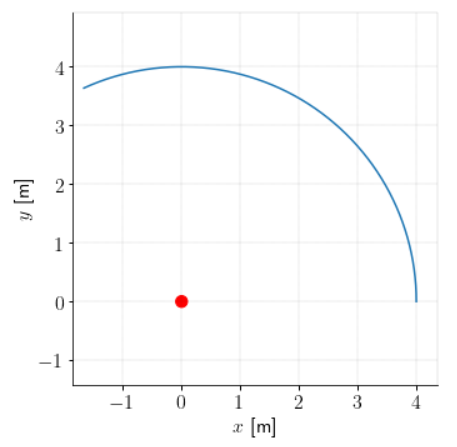}
    \caption{\textbf{Base Spacecraft Trajectory}}\label{fig:3}
  \end{minipage}
  \hfill
  \begin{minipage}[b]{0.4\textwidth}
    \includegraphics[width=\textwidth]{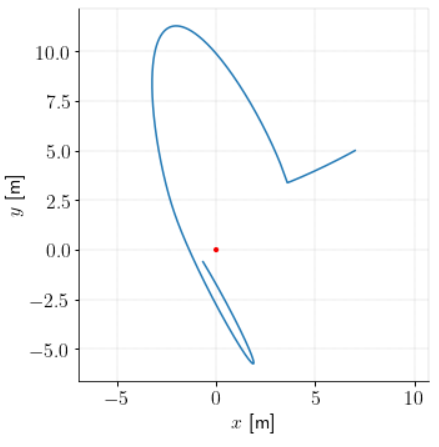}
    \caption{\textbf{Debris Trajectory}}\label{fig:4}
  \end{minipage}
\end{figure}

\noindent
Figure~\ref{fig:3} shows the path travelled by the base center of mass during the simulation interval, which corresponds to a portion of the circular orbit around the target (highlighted in red); as set, the trajectory is regular, not influenced by external dynamics. 
As shown in Figure~\ref{fig:4} the debris responds to collision with a sharp change in its direction, bouncing on the link surface and re-approaching following a circular path. Its physically realistic post-collision behavior contributes to validating the model reliability for this complex scenario, which requires handling collision resolution between multiple couples of bodies at a time.

\begin{figure}[!h]
  \centering
  \begin{minipage}[b]{0.4\textwidth}
    \includegraphics[width=\textwidth]{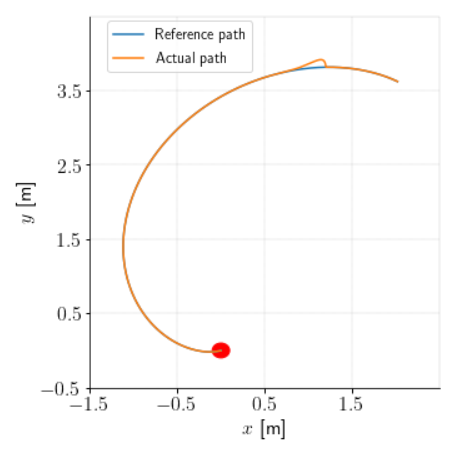}
    \caption{\textbf{End-Effector Trajectory}}\label{fig:5}
  \end{minipage}
  \hfill
  \begin{minipage}[b]{0.4\textwidth}
    \includegraphics[width=\textwidth]{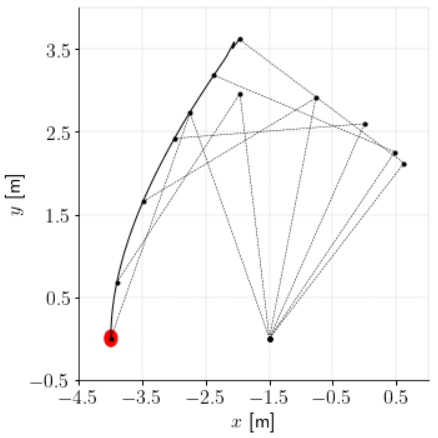}
    \caption{\textbf{Manipulator Configurations}}\label{fig:6}
  \end{minipage}
\end{figure}

\noindent
Figure~\ref{fig:5} compares the actual end-effector parabolic trajectory relative to the target with the reference one; despite a minor deviation caused by the impact, there is a subsequent exact overlap between the curves. 
Figure~\ref{fig:6} illustrates the trajectory of the end-effector in the base spacecraft reference system and the corresponding intermediate configurations of the manipulator.

\begin{figure}[!ht]
  \centering
  \begin{minipage}[b]{0.4\textwidth}
    \includegraphics[width=\textwidth]{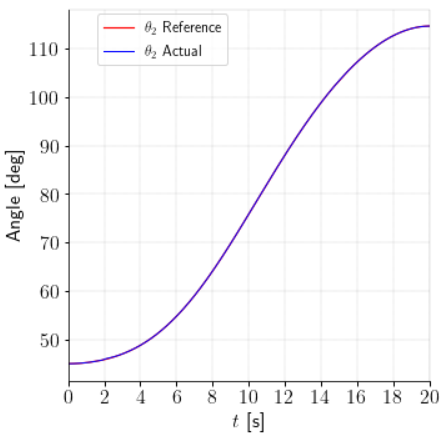}
    \caption{ \textbf{ $\boldsymbol{\theta_2}$ Trajectory  } }    \label{fig:7}
  \end{minipage}
  \hfill
  \begin{minipage}[b]{0.4\textwidth}
    \includegraphics[width=\textwidth]{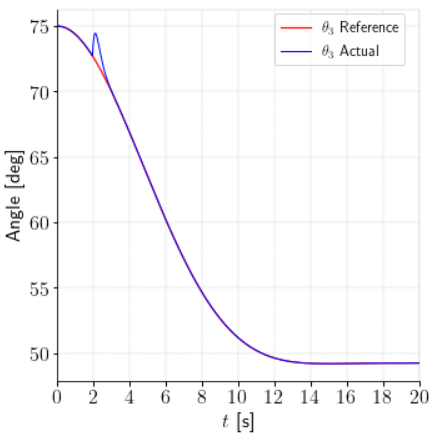}
    \caption{ \textbf{ $\boldsymbol{\theta_3}$ Trajectory  }   }  \label{fig:8}
  \end{minipage}
\end{figure}

\newpage
\noindent
Resulting trajectories for manipulator joints angles are shown in Figures~\ref{fig:7} and~\ref{fig:8}. 
At the impact instant, the manipulator experiences a sudden exchange of forces and momentum which greatly exceeds the driving torques and consequently cause a sudden deviation of the joints from the planned path. Each link is affected by the coupled inertia of the system and reacts accordingly; although the base is not influenced by the collision and prevents the manipulator from uncontrolled drift. 
The collision effect is especially noticeable in the joint of link II, which is directly hit by the debris.

\begin{figure}[!htbp]
  \centering
  \begin{minipage}[b]{0.4\textwidth}
    \includegraphics[width=\textwidth]{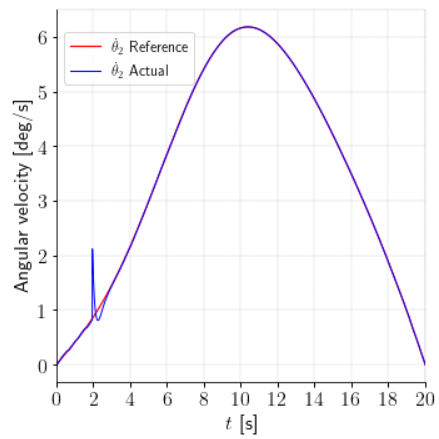}
    \caption{ \textbf{ $\boldsymbol{\theta_2}$ Angular Velocity  }}    \label{fig:9}
  \end{minipage}
  \hfill
  \begin{minipage}[b]{0.4\textwidth}
    \includegraphics[width=\textwidth]{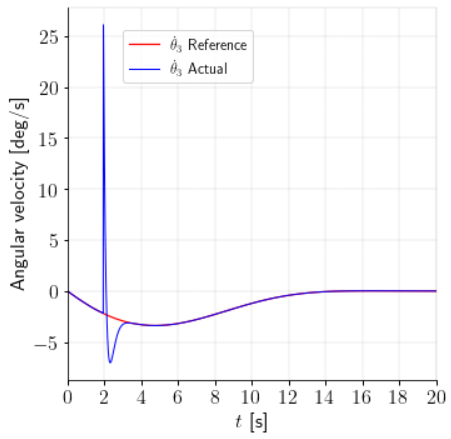}
    \caption{ \textbf{ $\boldsymbol{\theta_3}$  Angular Velocity  }}  \label{fig:10}
  \end{minipage}
\end{figure}

\begin{figure}[!h]
	\centering\includegraphics[width=3in]{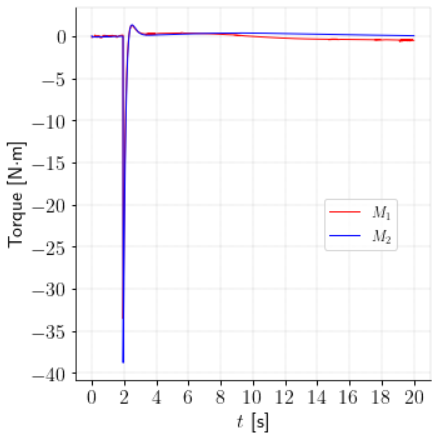}
	\caption{\textbf{Joint Torques}}
	\label{fig:11}
\end{figure}

\noindent
Figures~\ref{fig:9} and~\ref{fig:10} highlight a significant deviation of actual angular velocities from reference trajectories at the impact instant.
Nevertheless, the original motion is recovered and angular velocities of both links at instant t=20 sec decrease to 0, meaning that the relative end-effector velocity to the target at final time is 0, according to the requirements of soft-touch. 
\noindent
The subsequent increase of trajectory error is detected by the feedback part of the computed torque control, which prescribes the recovering maneuver immediately after the impact for re-stabilizing the original trajectory.
Except for the abrupt change caused by the collision, the driving torque functions imposed on motors are characterized by a smooth trend (Figure~\ref{fig:11}), reproducible in practice without exceeding saturation limits, even during the post-collision recovery maneuver.

\label{ch:Conclusion}
\section{Conclusions}
In this paper we presented a novel approach to simulate collisions in a dynamic system, which relies on modelling the contact dynamics in symbolic form, allowing for a direct and systematic integration in the equations of motion of the system. 
This approach was validated through extensive comparisons with the standard numerical approach, proving its reliability in obtaining consistent results and highlighting the benefits of an improved capability in modelling and analysing multibody systems.
We have demonstrated how the appropriate handling of the conversion between symbolic expressions and numerical counterparts can reduce calculus complexity and maximize evaluating speedup.  
The effective leveraging of both symbolic and numerical methods leads consequently to significant improvements in the system modelling and computational performances.

\bibliographystyle{AAS_publication}   % Number the references.
\bibliography{references}   % Use references.bib to resolve the labels.

\end{document}